%% file: iclr2023_conference.tex
\pdfoutput=1

\documentclass{article} 
\usepackage{iclr2023_conference,times}

\input{math_commands.tex}

\usepackage{hyperref}

\usepackage{url}
\usepackage{graphicx}
\title{Adapting to the Low-Resource Double-Bind: Investigating Low-Compute Methods on Low-resource African Languages}

\author{\normalsize $\forall$*, Colin Leong$^{1}$, Herumb Shandilya$^*$, Bonaventure F. P. Dossou $^{2,3,4,14}$, Atnafu Lambebo Tonja$^{5}$,\\ 
\textbf{\normalsize Joel Mathew$^{6}$,  Abdul-Hakeem Omotayo $^{7}$, Oreen Yousuf$^*$, Zainab Akinjobi $^{8}$, }\\
\textbf{\normalsize  Chris Chinenye Emezue $^{9,14}$, Shamsudeen Muhammad  $^{10}$, Steven Kolawole $^{11}$,}\\
\textbf{\normalsize Younwoo Choi $^{12}$, Tosin Adewumi $^{13}$} \\\\
\footnotesize
$^*$Masakhane NLP, $^1$University of Dayton, $^2$Center for Intelligent Machines, McGill University, $^3$Mila Quebec AI Institute,  \\ \footnotesize $^4$Lelapa AI, $^5$Instituto Politécnico Nacional,
$^6$USC Information Sciences Institute, $^7$University of California, Davis \\ \footnotesize $^8$New Mexico State University, $^9$ Technical University of Munich, $^{10}$University of Porto, $^{11}$ML Collective, 
\\ \footnotesize $^{12}$University of Toronto, $^{13}$ML Group, Luleå University of Technology,$^{14}$Lanfrica.
 \\
 \\
}

 \iclrfinalcopy 
\begin{document}

\maketitle
\maketitle
\makeatletter
\def\@fnsymbol#1{\ensuremath{\ifcase#1\or\forall\or\dagger\fi}}
\makeatother

\renewcommand{\thefootnote}{\fnsymbol{footnote}}
\footnotetext[1]{To represent the whole Masakhane community.}
\renewcommand{\thefootnote}{\arabic{footnote}}

\begin{abstract}
Many natural language processing (NLP) tasks make use of massively pre-trained language models, which are computationally expensive. However, access to high computational resources added to the issue of data scarcity of African languages constitutes a real barrier to research experiments on these languages. In this work, we explore the applicability of low-compute approaches such as language adapters in the context of this \textit{low-resource double-bind}. We intend to answer the following question: do language adapters allow those who are doubly bound by data and compute to practically build useful models? Through fine-tuning experiments on African languages, we evaluate their effectiveness as cost-effective approaches to low-resource African NLP. Using solely free compute resources, our results show that language adapters achieve comparable performances to massive pre-trained language models which are heavy on computational resources. This opens the door to further experimentation and exploration on full-extent of language adapters capacities.
\end{abstract}


\section{Introduction and Motivation: Adapting to The Low-resource Double-Bind Problem}

\cite{ahia-etal-2021-low-resource} coined the term \textit{low-resource double-bind} to describe the \textbf{co-occurrence of
data limitations and compute resource constraints}. Especially in the African setting, this double limitation often occurs  because most people do not have access to compute resources like GPUs and TPUs to construct different research projects that require more and more computational resources. The other limitation is the availability of datasets: African languages account for a small fraction
of available language resources, and NLP
research rarely considers them \citep{nekoto2020participatory}. This has a big impact on researchers working on different NLP tasks for African languages.  
In this study, we embrace this double limitation and investigate computationally efficient methods under low-data and compute conditions that will enable the researchers to work on different NLP tasks for African languages without being limited to the dataset and computational resources. We seek to answer if, and how, these can be used to build useful models for different NLP tasks. We focus specifically on training language and task adapters \cite{pfeiffer2020AdapterHub} and evaluating them on downstream Named Entity Recognition (NER) tasks.

\subsection{Unique Challenges for Double-Bind Model Training}
Training under the double-bind scenario introduces a number of unique challenges. For example, using free compute limits the size of the \textit{model}, \textit{dataset}; and it leads to the length of training. Resource limits may cause training runs to timeout, putting wall-clock limits on training.

\subsection{Language Adapters, Task Adapters and AdapterHub}
Fine-tuning all or a majority of a pre-trained model needs a lot of computational resources and also it depends on the availability of the dataset for the specific task which is the problem in the African context.
 \citet{pfeiffer2020AdapterHub} introduced AdapterHub, which is a central repository for pre-trained adapter modules. \textit{Adapters} refers to a set of newly introduced weights, typically within the layers of a transformer model. Adapters provide an alternative to fully fine-tuning the model for each downstream task while maintaining performance. Adapters also have the benefit of requiring as little as 1MB of storage space per task \citep{pmlr-v97-houlsby19a}. 
 Rather than pre-training a large language model, a \textit{language adapter} can be pre-trained for each language.
Adapter modules are parameter-efficient, sustainable, and achieve near SOTA results on low-resource and cross-lingual tasks \citep{pmlr-v97-houlsby19a, he-etal-2021-effectiveness}.
In this work, we leverage Adapter-based tuning for African languages because it has been shown to mitigate forgetting issues better than fine-tuning, as it yields representations with less deviation from those generated by the original pre-training \citep{he-etal-2021-effectiveness}.  Moreover, Adapter fine-tuning does not take a lot of time because of its lightweight nature: we believe this will be one of the main advantages for people who are highly limited with computational resources.








\section{Current Status and Results: Community Model Training Using Free Resources}

We ran a collaborative project, where community volunteers used free resources (i.e., Google Colab) to pre-train language adapters for several African languages, then used those language adapters to fine-tune on the MasakhaNER 1.0 and 2.0 datasets \citep{adelani-etal-2021-masakhaner, adelani-etal-2022-masakhaner}, to determine how much benefit the language adapters would provide. We used Weights and Biases \citep{wandb} for experiment tracking and analysis. Community Members were told not to use any computing resources that would not be available to a graduate student in Africa with limited funds. 

\subsection{Initial Experiments: Pre-train Language Adapters and Fine-tune on NER}

For our initial experiments, we concentrated on creating a pipeline to evaluate monolingual language adapters. We used default settings taken from AdapterHub examples, and pre-trained monolingual language-adapter modules using \textbf{\textit{roberta-base}} as the base model. Each language adapter was then used to perform downstream NER tasks on the respective language.
Each language adapter is pre-trained using the \textbf{\textit{MAFAND-MT}} dataset, the largest MT benchmark dataset for African languages in the news domain, covering 21 languages \citep{adelani-etal-2022-thousand}. We used target-language sentences to pre-train monolingual language adapters
, and finetuning on NER was performed using both MasakhaNER (1.0 and 2.0) datasets. To compare and evaluate the performance using language adapters in a downstream task, we conducted a baseline experiment by fine-tuning roberta-base pre-trained language model. Information about dataset splits has been presented and detailed in Table \ref{tab:lang-data}.

\begin{table}[h]
\small
\centering
\resizebox{\textwidth}{!}{%
\begin{tabular}{lllccccc}
\hline
Language (ISO) &
  Family &
  Region &
  \multicolumn{2}{l}{\textbf{Language Adapter Data}} &
  \multicolumn{3}{l}{\textbf{NER Finetuning Data}} \\ 
  &&&Train&Dev&Train&Dev&Test\\\hline
  
Amharic (amh)         & Afro-Asiatic-­Ethio-Semitic & East            & 1037 &  899                      & 1750   &250                     & 500                        \\
Fon (fon)             & Niger-Congo-Volta-Niger      & West            & 2637                         & 1227                        & 4343 & 621 & 1240                       \\
Hausa (hau)           & Afro-Asiatic-Chadic        & West    & 5865  &1300 & 1903&2072 & 545 \\
Igbo (ibo)            & Niger-Congo-Volta-Niger      & West            & 6998 &1500                        & 2233 &319& 638 \\
Kinyarwanda (kin)     & Niger-Congo-Bantu            & East & 1006  & 460&2110&301&604 \\
Luganda (lug)         & Niger-Congo-Bantu            & East            & 4075                         &1500& 2003   &200                     & 401                        \\
Nigerian-Pidgin (pcm) & English Creole               & West& 4790 & 1484& 2100&300&600                        \\
Swahili (swa)& Niger-Congo-Bantu & East \& Central & 30782 & 1791 &2104 &300& 602 \\
Akan/Twi (twi)        & Niger-Congo-Kwa              & West& 3337 &  1284           & 4240                        &605& 1211                       \\
Wolof (wol)           & Niger-Congo-Senegambia       & West & 3360  &1506& 1871 & 267&536 \\
Yorùbá (yor)          & Niger-Congo-Volta-Niger      & West & 6644  &1544& 2124&303                        & 608                        \\
Zulu (zul)            & Niger-Congo-Bantu            & South           & 3500 &1239                        & 5848     &836                   & 1670                       \\ \hline
\end{tabular}%
} 
\caption{Languages with ISO 639-2 Code. Language adapter training data was taken from the MAFAND dataset. NER fine-tuning and Evaluation data was taken from the MasakhaNER and MasakhaNER 2.0 datasets.}
\label{tab:lang-data}
\end{table}




\begin{table*}[h]
\centering
\begin{tabular}{lcccc}
\hline
\textbf{Language}  &\multicolumn{4}{c}{\textbf{F1 - Score}}   \\ 
&\multicolumn{2}{l}{Baseline NER}  &\multicolumn{2}{l}{Adapter NER}\\
& Dev & Test & Dev & Test \\ \hline
Amharic     &0.32&\textbf{0.34}    & 0.29                         & 0.27                            \\
Fon           &0.83&0.79  & 0.82&\textbf{0.80}  \\
Hausa         &0.90&\textbf{0.85}     & 0.85                        & 0.79                            \\
Igbo           &0.84&\textbf{0.79}    & 0.65  & 0.69                              \\
Kinyarwanda    &0.79&\textbf{0.68}    & 0.64  &0.60\\
Luganda        &0.67&\textbf{0.73}  & 0.64 &0.70 \\
Nigerian-Pidgin   &0.90&\textbf{0.87}& 0.89  &0.83\\
Swahili        &0.84&\textbf{0.81}   & 0.81  & 0.78     \\
Akan/Twi      &0.77&\textbf{0.75}   & 0.75   & 0.73 \\
Wolof        &0.70&\textbf{0.57}     &0.68    & 0.56    \\
Yorùbá  &0.66&\textbf{0.71} & 0.65 & 0.68    \\
Zulu       &0.78&\textbf{0.83}   & 0.76    & 0.80                             \\ \hline
Average       &0.75&\textbf{0.72}   & 0.72    & \textit{\underline{0.69}}                             \\ \hline
\end{tabular}%

\caption{Results for averaged eval and predict F1 scores by language.}
\label{tab:results}
\end{table*}



\section{Results and Future works}
In this section, we discuss the experimental results of our approach and future works.
\subsection{Adapters Enable Rapid Iteration}
In Table \ref{tab:results}, we present the results of two experiments, across 12 African Languages:
\begin{itemize}
    \item Baseline NER: setup where \textit{roberta-base} has been used to directly perform NER downstream (finetuning and evaluation) using MasakhaNER 1.0 and 2.0.
    \item Adapter NER: setup where we first of all trained a language adapter based on \textit{roberta-base}, then used the latter to perform downstream NER task.
\end{itemize}

Our initial results (Table \ref{tab:results}) with language adapters show comparable average performance to \textit{roberta-base} finetuned on the NER downstream task: this demonstrates that it is indeed feasible to train a monolingual language adapter in African low-resource settings, only with free computational resources while achieving comparable performance to massive pre-trained language model which requires a lot of computational resources.

Given how rapidly adapters can be trained and re-trained, it is possible to conduct rapid and iterative experiments. Therefore, there are many experiments we further wish to explore:
\begin{itemize}
    \item Extending experiments to Other Base Models: We will extend our experiments to several massive multilingual pre-trained language models like XLM-R \citep{conneau-etal-2020-unsupervised}, but also to Afro-centric language models like AfroLM \citep{dossou-etal-2022-afrolm}, AfriBERTa \citep{ogueji-etal-2021-small}, and AfroXLMR \citep{alabi-etal-2022-adapting}. This will allow direct comparison with previous benchmarks on MasakhaNER datasets.
    \item Alleviating Low-Data Issues with Phylogeny-based Methods: We will analyze how phylogeny-based adapter training affects performance. \citet{faisal-anastasopoulos-2022-phylogeny} carried this task out on other, non-African, low-resource languages. As training individual language adapters have proved to be time-efficient, we hope that training adapters on multiple, linguistically similar, languages yields better results.
    \item Quantifying Performance Improvement Tradeoffs vs Dataset Size: Given how quickly adapters can be trained, we can determine ratios of optimal data size to maximize performance. Power usage would have to be analyzed in this context as well.
    \item Other low-compute methods: In addition to the AdapterHub paradigm, we wish to comparatively analyze other low-compute methods such as \citet{cramming_https://doi.org/10.48550/arxiv.2212.14034}
\end{itemize} 

\section{Conclusion}
We built a pipeline for analyzing low-compute methods on low-resource languages. Initial results suggest that language adapter modules can be quickly and easily trained on entirely free resources such as Google Colab, opening the door to further experimentation and exploration. We hope to conduct further rounds of experiments and release both trained models and best practices for practical training of models when constrained by both low-data and low-compute.

\bibliography{iclr2023_conference}
\bibliographystyle{iclr2023_conference}


\end{document}

%% file: math_commands.tex
\usepackage{amsmath,amsfonts,bm}

\def\eqref#1{equation~\ref{#1}}

\def\1{\bm{1}}

\DeclareMathAlphabet{\mathsfit}{\encodingdefault}{\sfdefault}{m}{sl}
\SetMathAlphabet{\mathsfit}{bold}{\encodingdefault}{\sfdefault}{bx}{n}













%% file: iclr2023_conference.bbl
\begin{thebibliography}{15}
\providecommand{\natexlab}[1]{#1}
\providecommand{\url}[1]{\texttt{#1}}
\expandafter\ifx\csname urlstyle\endcsname\relax
  \providecommand{\doi}[1]{doi: #1}\else
  \providecommand{\doi}{doi: \begingroup \urlstyle{rm}\Url}\fi

\bibitem[Adelani et~al.(2022{\natexlab{a}})Adelani, Alabi, Fan, Kreutzer, Shen,
  Reid, Ruiter, Klakow, Nabende, Chang, Gwadabe, Sackey, Dossou, Emezue, Leong,
  Beukman, Muhammad, Jarso, Yousuf, Niyongabo~Rubungo, Hacheme, Wairagala,
  Nasir, Ajibade, Ajayi, Gitau, Abbott, Ahmed, Ochieng, Aremu, Ogayo, Mukiibi,
  Ouoba~Kabore, Kalipe, Mbaye, Tapo, Memdjokam~Koagne, Munkoh-Buabeng, Wagner,
  Abdulmumin, Awokoya, Buzaaba, Sibanda, Bukula, and
  Manthalu]{adelani-etal-2022-thousand}
David Adelani, Jesujoba Alabi, Angela Fan, Julia Kreutzer, Xiaoyu Shen, Machel
  Reid, Dana Ruiter, Dietrich Klakow, Peter Nabende, Ernie Chang, Tajuddeen
  Gwadabe, Freshia Sackey, Bonaventure F.~P. Dossou, Chris Emezue, Colin Leong,
  Michael Beukman, Shamsuddeen Muhammad, Guyo Jarso, Oreen Yousuf, Andre
  Niyongabo~Rubungo, Gilles Hacheme, Eric~Peter Wairagala, Muhammad~Umair
  Nasir, Benjamin Ajibade, Tunde Ajayi, Yvonne Gitau, Jade Abbott, Mohamed
  Ahmed, Millicent Ochieng, Anuoluwapo Aremu, Perez Ogayo, Jonathan Mukiibi,
  Fatoumata Ouoba~Kabore, Godson Kalipe, Derguene Mbaye, Allahsera~Auguste
  Tapo, Victoire Memdjokam~Koagne, Edwin Munkoh-Buabeng, Valencia Wagner, Idris
  Abdulmumin, Ayodele Awokoya, Happy Buzaaba, Blessing Sibanda, Andiswa Bukula,
  and Sam Manthalu.
\newblock A few thousand translations go a long way! leveraging pre-trained
  models for {A}frican news translation.
\newblock In \emph{Proceedings of the 2022 Conference of the North American
  Chapter of the Association for Computational Linguistics: Human Language
  Technologies}, pp.\  3053--3070, Seattle, United States, July
  2022{\natexlab{a}}. Association for Computational Linguistics.
\newblock \doi{10.18653/v1/2022.naacl-main.223}.
\newblock URL \url{https://aclanthology.org/2022.naacl-main.223}.

\bibitem[Adelani et~al.(2022{\natexlab{b}})Adelani, Neubig, Ruder, Rijhwani,
  Beukman, Palen-Michel, Lignos, Alabi, Muhammad, Nabende, Dione, Bukula,
  Mabuya, Dossou, Sibanda, Buzaaba, Mukiibi, Kalipe, Mbaye, Taylor, Kabore,
  Emezue, Aremu, Ogayo, Gitau, Munkoh-Buabeng, Memdjokam~Koagne, Tapo, Macucwa,
  Marivate, Elvis, Gwadabe, Adewumi, Ahia, Nakatumba-Nabende, Mokono, Ezeani,
  Chukwuneke, Oluwaseun~Adeyemi, Hacheme, Abdulmumin, Ogundepo, Yousuf, Moteu,
  and Klakow]{adelani-etal-2022-masakhaner}
David Adelani, Graham Neubig, Sebastian Ruder, Shruti Rijhwani, Michael
  Beukman, Chester Palen-Michel, Constantine Lignos, Jesujoba Alabi,
  Shamsuddeen Muhammad, Peter Nabende, Cheikh M.~Bamba Dione, Andiswa Bukula,
  Rooweither Mabuya, Bonaventure F.~P. Dossou, Blessing Sibanda, Happy Buzaaba,
  Jonathan Mukiibi, Godson Kalipe, Derguene Mbaye, Amelia Taylor, Fatoumata
  Kabore, Chris~Chinenye Emezue, Anuoluwapo Aremu, Perez Ogayo, Catherine
  Gitau, Edwin Munkoh-Buabeng, Victoire Memdjokam~Koagne, Allahsera~Auguste
  Tapo, Tebogo Macucwa, Vukosi Marivate, Mboning~Tchiaze Elvis, Tajuddeen
  Gwadabe, Tosin Adewumi, Orevaoghene Ahia, Joyce Nakatumba-Nabende, Neo~Lerato
  Mokono, Ignatius Ezeani, Chiamaka Chukwuneke, Mofetoluwa Oluwaseun~Adeyemi,
  Gilles~Quentin Hacheme, Idris Abdulmumin, Odunayo Ogundepo, Oreen Yousuf,
  Tatiana Moteu, and Dietrich Klakow.
\newblock {M}asakha{NER} 2.0: {A}frica-centric transfer learning for named
  entity recognition.
\newblock In \emph{Proceedings of the 2022 Conference on Empirical Methods in
  Natural Language Processing}, pp.\  4488--4508, Abu Dhabi, United Arab
  Emirates, December 2022{\natexlab{b}}. Association for Computational
  Linguistics.
\newblock URL \url{https://aclanthology.org/2022.emnlp-main.298}.

\bibitem[Adelani et~al.(2021)Adelani, Abbott, Neubig, D{'}souza, Kreutzer,
  Lignos, Palen-Michel, Buzaaba, Rijhwani, Ruder, Mayhew, Azime, Muhammad,
  Emezue, Nakatumba-Nabende, Ogayo, Anuoluwapo, Gitau, Mbaye, Alabi, Yimam,
  Gwadabe, Ezeani, Niyongabo, Mukiibi, Otiende, Orife, David, Ngom, Adewumi,
  Rayson, Adeyemi, Muriuki, Anebi, Chukwuneke, Odu, Wairagala, Oyerinde, Siro,
  Bateesa, Oloyede, Wambui, Akinode, Nabagereka, Katusiime, Awokoya, MBOUP,
  Gebreyohannes, Tilaye, Nwaike, Wolde, Faye, Sibanda, Ahia, Dossou, Ogueji,
  DIOP, Diallo, Akinfaderin, Marengereke, and
  Osei]{adelani-etal-2021-masakhaner}
David~Ifeoluwa Adelani, Jade Abbott, Graham Neubig, Daniel D{'}souza, Julia
  Kreutzer, Constantine Lignos, Chester Palen-Michel, Happy Buzaaba, Shruti
  Rijhwani, Sebastian Ruder, Stephen Mayhew, Israel~Abebe Azime, Shamsuddeen~H.
  Muhammad, Chris~Chinenye Emezue, Joyce Nakatumba-Nabende, Perez Ogayo, Aremu
  Anuoluwapo, Catherine Gitau, Derguene Mbaye, Jesujoba Alabi, Seid~Muhie
  Yimam, Tajuddeen~Rabiu Gwadabe, Ignatius Ezeani, Rubungo~Andre Niyongabo,
  Jonathan Mukiibi, Verrah Otiende, Iroro Orife, Davis David, Samba Ngom, Tosin
  Adewumi, Paul Rayson, Mofetoluwa Adeyemi, Gerald Muriuki, Emmanuel Anebi,
  Chiamaka Chukwuneke, Nkiruka Odu, Eric~Peter Wairagala, Samuel Oyerinde,
  Clemencia Siro, Tobius~Saul Bateesa, Temilola Oloyede, Yvonne Wambui, Victor
  Akinode, Deborah Nabagereka, Maurice Katusiime, Ayodele Awokoya, Mouhamadane
  MBOUP, Dibora Gebreyohannes, Henok Tilaye, Kelechi Nwaike, Degaga Wolde,
  Abdoulaye Faye, Blessing Sibanda, Orevaoghene Ahia, Bonaventure F.~P. Dossou,
  Kelechi Ogueji, Thierno~Ibrahima DIOP, Abdoulaye Diallo, Adewale Akinfaderin,
  Tendai Marengereke, and Salomey Osei.
\newblock {M}asakha{NER}: Named entity recognition for {A}frican languages.
\newblock \emph{Transactions of the Association for Computational Linguistics},
  9:\penalty0 1116--1131, 2021.
\newblock \doi{10.1162/tacl_a_00416}.
\newblock URL \url{https://aclanthology.org/2021.tacl-1.66}.

\bibitem[Ahia et~al.(2021)Ahia, Kreutzer, and
  Hooker]{ahia-etal-2021-low-resource}
Orevaoghene Ahia, Julia Kreutzer, and Sara Hooker.
\newblock The low-resource double bind: An empirical study of pruning for
  low-resource machine translation.
\newblock In \emph{Findings of the Association for Computational Linguistics:
  EMNLP 2021}, pp.\  3316--3333, Punta Cana, Dominican Republic, November 2021.
  Association for Computational Linguistics.
\newblock \doi{10.18653/v1/2021.findings-emnlp.282}.
\newblock URL \url{https://aclanthology.org/2021.findings-emnlp.282}.

\bibitem[Alabi et~al.(2022)Alabi, Adelani, Mosbach, and
  Klakow]{alabi-etal-2022-adapting}
Jesujoba~O. Alabi, David~Ifeoluwa Adelani, Marius Mosbach, and Dietrich Klakow.
\newblock Adapting pre-trained language models to {A}frican languages via
  multilingual adaptive fine-tuning.
\newblock In \emph{Proceedings of the 29th International Conference on
  Computational Linguistics}, pp.\  4336--4349, Gyeongju, Republic of Korea,
  October 2022. International Committee on Computational Linguistics.
\newblock URL \url{https://aclanthology.org/2022.coling-1.382}.

\bibitem[Biewald(2020)]{wandb}
Lukas Biewald.
\newblock Experiment tracking with weights and biases, 2020.
\newblock URL \url{https://www.wandb.com/}.
\newblock Software available from wandb.com.

\bibitem[Conneau et~al.(2020)Conneau, Khandelwal, Goyal, Chaudhary, Wenzek,
  Guzm{\'a}n, Grave, Ott, Zettlemoyer, and
  Stoyanov]{conneau-etal-2020-unsupervised}
Alexis Conneau, Kartikay Khandelwal, Naman Goyal, Vishrav Chaudhary, Guillaume
  Wenzek, Francisco Guzm{\'a}n, Edouard Grave, Myle Ott, Luke Zettlemoyer, and
  Veselin Stoyanov.
\newblock Unsupervised cross-lingual representation learning at scale.
\newblock In \emph{Proceedings of the 58th Annual Meeting of the Association
  for Computational Linguistics}, pp.\  8440--8451, Online, July 2020.
  Association for Computational Linguistics.
\newblock \doi{10.18653/v1/2020.acl-main.747}.
\newblock URL \url{https://aclanthology.org/2020.acl-main.747}.

\bibitem[Dossou et~al.(2022)Dossou, Tonja, Yousuf, Osei, Oppong, Shode,
  Awoyomi, and Emezue]{dossou-etal-2022-afrolm}
Bonaventure F.~P. Dossou, Atnafu Tonja, Oreen Yousuf, Salomey Osei, Abigail
  Oppong, Iyanuoluwa Shode, Oluwabusayo~Olufunke Awoyomi, and Chris Emezue.
\newblock {A}fro{LM}: A self-active learning-based multilingual pretrained
  language model for 23 {A}frican languages.
\newblock In \emph{Proceedings of The Third Workshop on Simple and Efficient
  Natural Language Processing (SustaiNLP)}, pp.\  52--64, Abu Dhabi, United
  Arab Emirates (Hybrid), December 2022. Association for Computational
  Linguistics.
\newblock URL \url{https://aclanthology.org/2022.sustainlp-1.11}.

\bibitem[Faisal \& Anastasopoulos(2022)Faisal and
  Anastasopoulos]{faisal-anastasopoulos-2022-phylogeny}
Fahim Faisal and Antonios Anastasopoulos.
\newblock Phylogeny-inspired adaptation of multilingual models to new
  languages.
\newblock In \emph{Proceedings of the 2nd Conference of the Asia-Pacific
  Chapter of the Association for Computational Linguistics and the 12th
  International Joint Conference on Natural Language Processing (Volume 1: Long
  Papers)}, pp.\  434--452, Online only, November 2022. Association for
  Computational Linguistics.
\newblock URL \url{https://aclanthology.org/2022.aacl-main.34}.

\bibitem[Geiping \& Goldstein(2022)Geiping and
  Goldstein]{cramming_https://doi.org/10.48550/arxiv.2212.14034}
Jonas Geiping and Tom Goldstein.
\newblock Cramming: Training a language model on a single gpu in one day, 2022.
\newblock URL \url{https://arxiv.org/abs/2212.14034}.

\bibitem[He et~al.(2021)He, Liu, Ye, Tan, Ding, Cheng, Low, Bing, and
  Si]{he-etal-2021-effectiveness}
Ruidan He, Linlin Liu, Hai Ye, Qingyu Tan, Bosheng Ding, Liying Cheng, Jiawei
  Low, Lidong Bing, and Luo Si.
\newblock On the effectiveness of adapter-based tuning for pretrained language
  model adaptation.
\newblock In \emph{Proceedings of the 59th Annual Meeting of the Association
  for Computational Linguistics and the 11th International Joint Conference on
  Natural Language Processing (Volume 1: Long Papers)}, pp.\  2208--2222,
  Online, August 2021. Association for Computational Linguistics.
\newblock \doi{10.18653/v1/2021.acl-long.172}.
\newblock URL \url{https://aclanthology.org/2021.acl-long.172}.

\bibitem[Houlsby et~al.(2019)Houlsby, Giurgiu, Jastrzebski, Morrone,
  De~Laroussilhe, Gesmundo, Attariyan, and Gelly]{pmlr-v97-houlsby19a}
Neil Houlsby, Andrei Giurgiu, Stanislaw Jastrzebski, Bruna Morrone, Quentin
  De~Laroussilhe, Andrea Gesmundo, Mona Attariyan, and Sylvain Gelly.
\newblock Parameter-efficient transfer learning for {NLP}.
\newblock In Kamalika Chaudhuri and Ruslan Salakhutdinov (eds.),
  \emph{Proceedings of the 36th International Conference on Machine Learning},
  volume~97 of \emph{Proceedings of Machine Learning Research}, pp.\
  2790--2799. PMLR, 09--15 Jun 2019.
\newblock URL \url{https://proceedings.mlr.press/v97/houlsby19a.html}.

\bibitem[Nekoto et~al.(2020)Nekoto, Marivate, Matsila, Fasubaa, Kolawole,
  Fagbohungbe, Akinola, Muhammad, Kabongo, Osei,
  et~al.]{nekoto2020participatory}
Wilhelmina Nekoto, Vukosi Marivate, Tshinondiwa Matsila, Timi Fasubaa, Tajudeen
  Kolawole, Taiwo Fagbohungbe, Solomon~Oluwole Akinola, Shamsuddeen~Hassan
  Muhammad, Salomon Kabongo, Salomey Osei, et~al.
\newblock Participatory research for low-resourced machine translation: A case
  study in african languages.
\newblock \emph{arXiv preprint arXiv:2010.02353}, 2020.

\bibitem[Ogueji et~al.(2021)Ogueji, Zhu, and Lin]{ogueji-etal-2021-small}
Kelechi Ogueji, Yuxin Zhu, and Jimmy Lin.
\newblock Small data? no problem! exploring the viability of pretrained
  multilingual language models for low-resourced languages.
\newblock In \emph{Proceedings of the 1st Workshop on Multilingual
  Representation Learning}, pp.\  116--126, Punta Cana, Dominican Republic,
  November 2021. Association for Computational Linguistics.
\newblock \doi{10.18653/v1/2021.mrl-1.11}.
\newblock URL \url{https://aclanthology.org/2021.mrl-1.11}.

\bibitem[Pfeiffer et~al.(2020)Pfeiffer, R\"uckl\'{e}, Poth, Kamath, Vuli\'{c},
  Ruder, Cho, and Gurevych]{pfeiffer2020AdapterHub}
Jonas Pfeiffer, Andreas R\"uckl\'{e}, Clifton Poth, Aishwarya Kamath, Ivan
  Vuli\'{c}, Sebastian Ruder, Kyunghyun Cho, and Iryna Gurevych.
\newblock Adapterhub: A framework for adapting transformers.
\newblock In \emph{Proceedings of the 2020 Conference on Empirical Methods in
  Natural Language Processing (EMNLP 2020): Systems Demonstrations}, pp.\
  46--54, Online, 2020. Association for Computational Linguistics.
\newblock URL \url{https://www.aclweb.org/anthology/2020.emnlp-demos.7}.

\end{thebibliography}
